\documentclass[runningheads]{llncs}

\usepackage[utf8]{inputenc}
\usepackage{graphicx}
\usepackage{amsmath,amssymb}
\usepackage{array}
\usepackage{color,soul}
\usepackage{algorithm}
\usepackage{algpseudocode}
\usepackage{siunitx}
\usepackage{booktabs}
\usepackage{svg}
\usepackage{bbm}
\usepackage{multirow}

\usepackage[font={footnotesize}, labelfont=bf, labelsep=period]{caption}
\newif\ifdraft
\drafttrue

\definecolor{orange}{rgb}{1,0.5,0}
\definecolor{pink}{rgb}{0.98, 0.38, 0.5}

\ifdraft
 \newcommand{\RS}[1]{{\color{red}{\bf RS: #1}}}
 
 \newcommand{\PMN}[1]{{\color{orange}{\bf PMN: #1}}}

 \newcommand{\MH}[1]{{\color{pink}{\bf MH: #1}}}

\else
 \newcommand{\RS}[1]{{\color{red}{}}}
 
 \newcommand{\PMN}[1]{{\color{red}{}}}
 
  \newcommand{\MH}[1]{{\color{red}{}}}
 
\fi

\newcommand{\real}{\mathbb{R}}
\newcommand{\x}{\mathbf{x}}
\newcommand{\y}{\mathbf{y}}
\newcommand{\z}{\mathbf{z}}

\newcommand{\V}{\mathbf{V}}
\renewcommand{\v}{\mathbf{v}}
\newcommand{\W}{\mathbf{W}}

\renewcommand{\b}{\mathbf{b}}

\DeclareMathOperator{\tr}{tr}
\DeclareMathOperator*{\argmin}{arg\,min}

\newcommand{\comment}[1]{}

\newcommand{\ie}{i.\,e.,\ }

\begin{document}
\title{Deep Multi-Label Classification in \\ Affine Subspaces}

\titlerunning{Deep Multi Label Classification in Affine Subspaces}
\author{Thomas Kurmann\inst{1} \and Pablo Márquez Neila\inst{1} \and Sebastian Wolf\inst{2} \and Raphael Sznitman\inst{1}}
\authorrunning{T. Kurmann et al.}
% index{Kurmann, Thomas}
% index{Márquez Neila, Pablo}
% index{Wolf, Sebastian}
% index{Sznitman, Raphael}

\institute{
University of Bern, Bern, Switzerland
\and
University Hospital of Bern, Bern,  Switzerland
}
\maketitle              % typeset the header of the contribution
\begin{abstract}
% !TEX root = top.tex
% !TEX spellcheck = en-US
Multi-label classification (MLC) problems are becoming increasingly popular in the context of medical imaging.
This has in part been driven by the fact that acquiring annotations for MLC is far less burdensome than for semantic segmentation and yet provides more expressiveness than multi-class classification.
However, to train MLCs, most methods have resorted to similar objective functions as with traditional multi-class classification settings.
We show in this work that such approaches are not optimal and instead propose a novel deep MLC
classification method in affine subspace.
At its core, the method attempts to pull features of class-labels towards different affine subspaces while maximizing the distance between them.
We evaluate the method using two MLC medical imaging datasets and show a large performance increase compared to previous multi-label frameworks.
This method can be seen as a plug-in replacement loss function and is trainable in an end-to-end fashion.

\end{abstract}

% !TEX root = top.tex
% !TEX spellcheck = en-US

\section{Introduction}
\label{sec:intro}

% Multi-label problems in MIC
In recent years, multi-label classification (MLC) tasks have gained important relevance in
medical imaging. In essence, MLC is focused on training prediction functions that can tag
image data with multiple labels that are not necessarily mutually
exclusive~\cite{Gibaja2014}. With the advent of Deep Learning, impressive performance in
the area of chest X-ray tagging~\cite{Wang2017-zo}, identification of disease comorbidity~\cite{Adeli2018} and retinal image characterization~\cite{De_Fauw2018-gk} have already been shown.

\begin{figure}[!t]
  \centering
  \includegraphics[width=0.49\textwidth]{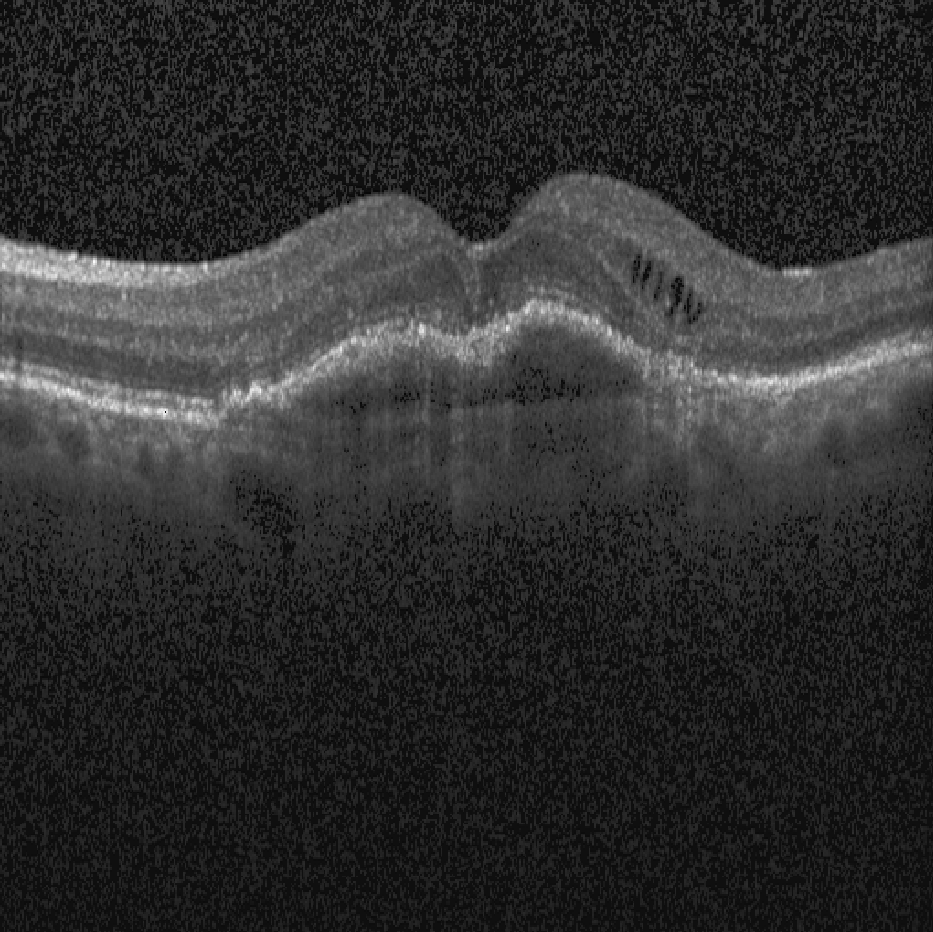}
  \includegraphics[width=0.49\textwidth]{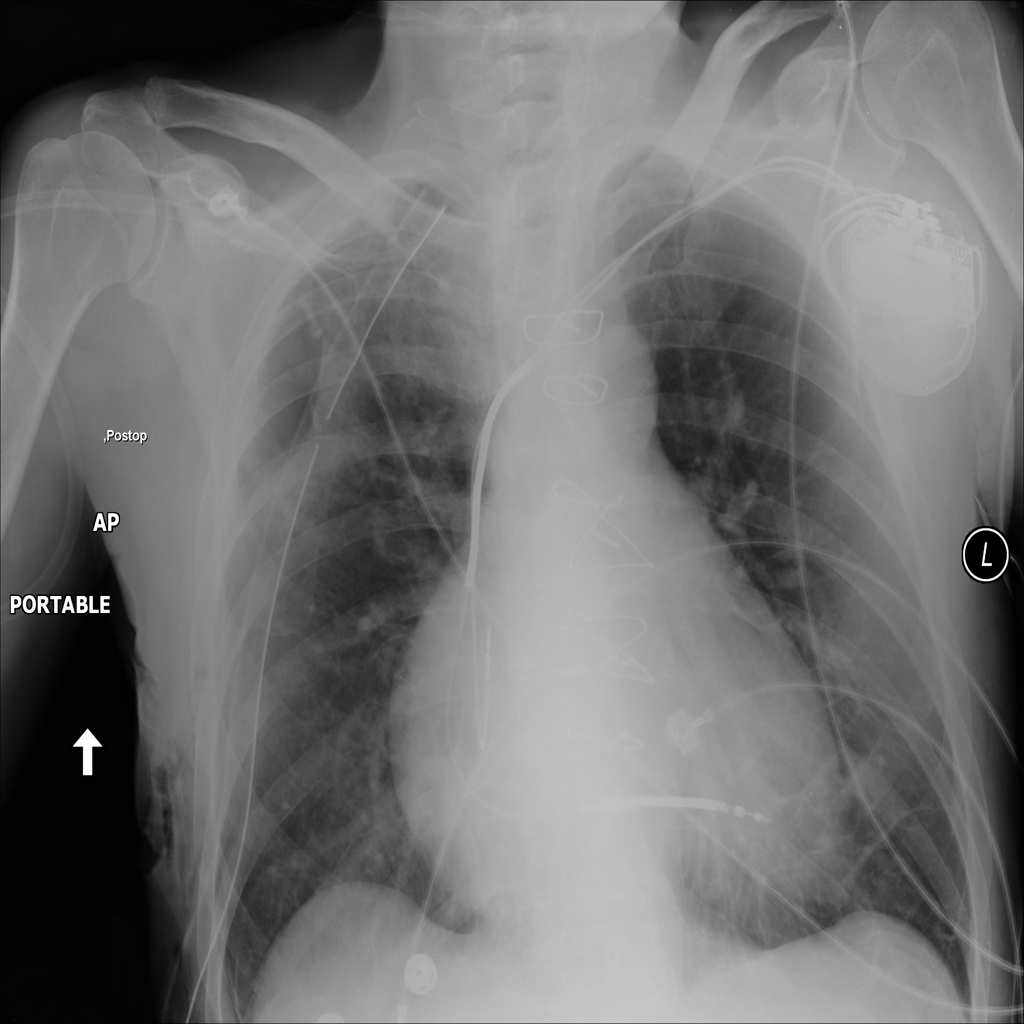}
\caption{\textbf{Left:} Retinal Optical Coherence Tomography scan containing 11 possible biomarkers. In this example Intraretinal Cysts and Fibrovascular PED are present. \textbf{Right:} Chest X-Ray scan~\cite{Wang2017-zo} with Cardiomegaly and Emphysema present.
}
\label{fig:examples}
\end{figure}

% Existing approaches
Yet, at its core, MLC is challenging because of the large number of possible label combinations a method needs to be able to predict and that some output configurations may be extremely rare even though they are important. For instance, for a MLC task with 11 binary labels as depicted in Fig.~\ref{fig:examples}, the number of possible prediction configuration outcomes is $2^{11}$. Given that generating groundtruth annotations for large dataset is often both time consuming and expensive, it is thus common for MLC training sets to not have any examples for many label combinations.

To overcome this, MLC problems were initially treated as direct extensions of binary
classification tasks, whereby multiple binary classifiers were trained independently to
predict each label~\cite{Read2011-rb}. In an effort to leverage common image features, recent
approaches have looked to share information across different labeling tasks using deep neural networks~\cite{Wang2017-zo}. In particular, while feature sharing is achieved via network weights, classification boundaries are optimized using traditional binary multi-class objectives.
An alternative is to consider that features associated to specific labels should be similar to each other, so to group these together in a subspace and be modeled by density approximators such as k-Nearest Neighbor (kNN)~\cite{Zhang2007-cu}.
Similarly, ranking loss functions have been used to disentagle labels in the feature space~\cite{Li2017-if}.
Recently, methods performing MLC in latent embedding spaces have been proposed for low dimensional input spaces which are not trainable end-to-end \cite{Li2015-uf} or require multiple networks \cite{Yeh2017-yl}.

% Unfortunately, a limitation of this, is that comparisons occur at sample pairs and does not scale well for non-disjoint labels.
% Similarly,~\cite{Caron2018-ky} considers the MLC in a multi-class classification setting by using centroid clustering and the cross entropy loss to avoid the centroids from collapsing.

To overcome these shortcomings, we present a novel framework for MLC.
Unlike most traditional approaches that attempt to discriminate training samples via decision boundaries that separate different labels, our approach is to enforce that samples with the same label values lie on a dedicated subspace.
To do this, we introduce a novel loss function that on one hand enforces that samples with the same label value are close to the same subspace, and on the other, that different subspaces are far apart from each other.
As such, when training a neural network (NN) with our approach, samples are pulled towards the learned subspaces and can easily be classified by means of a density estimation approach. To show this, we validated our approach on two MLC tasks (\ie OCT biomarker classification and chest X-ray tagging) using common NN architectures.
We show that our approach provides superior performances to a number of state-of-the-art methods for the same task.

% !TEX root = top.tex
% !TEX spellcheck = en-US

\section{Method}
\label{sec:method}
In a MLC, each input image $\x \in \real^{h \times w \times c}$ has $n$~different binary labels $\y = (y_0, \ldots, y_n)$ with $y_i\in\{0, 1\}$. The goal is finding a deep network~$f:\real^{h \times w \times c} \to [0, 1]^n$ such that $f(\x)_i$ is the estimated probability that label $y_i$~is~$1$ for the input image~$\x$. For convenience, we express our deep network as the composition of two functions~$f=g\circ{}h$: the \emph{feature extraction} function~$h:\real^{h \times w \times c} \to \real^d$ builds a $d$-dimensional descriptor vector for the given image~$\x$, and $g:\real^d\to [0, 1]^n$ is a multi-output binary classifier. Typically, the choice for $g$ is the standard multi-output logistic regression,

\begin{figure}[!b]
\centering
\includegraphics[width=0.99\textwidth]{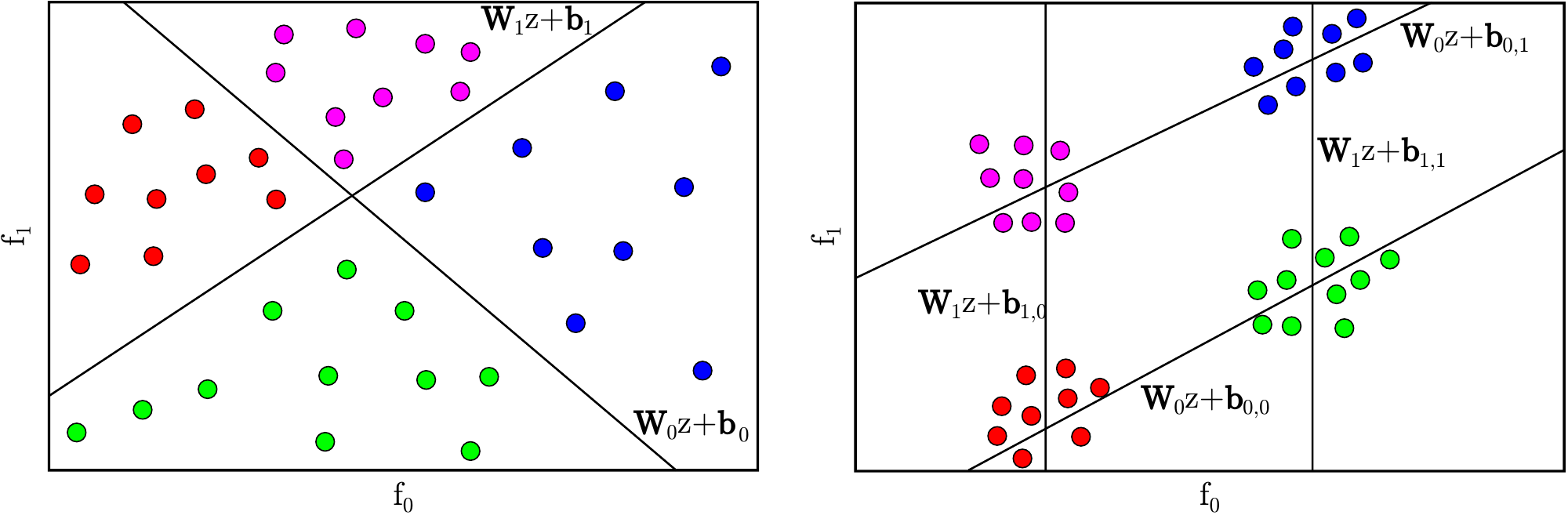}
\caption{Illustration of MLC (left) and our proposed \textbf{AS-MLC} method (right). This synthetic example the feature space is of size $d=2$ with $n=2$ labels: Red points = (00), Green = (01), Pink = (10) and Blue = (11). Note how the joint distribution of labels is clustered at the intersections of the hyperplanes.
}
\label{fig:hyperplanes}
\end{figure}

\begin{equation}
    g(\z) = \sigma(\V\z + \v),
\end{equation}
where $\V\in\real^{n\times{} d}$ and $\v\in\real^n$ define an affine transformation mapping from the feature space to $\real^n$, and $\sigma$~is the element-wise logistic function that provides final probabilities. In this case, the logistic regression splits the feature space using $n$~different $(d-1)$-dimensional hyperplanes, one for each label, and pushes each sample towards one side of each hyperplane depending on its labels. That is, it defines $2^n$~disjoint regions in the feature space (assuming that~$d\geq n$), one for each possible combination of labels, and moves samples to their corresponding regions as shown in Fig.~\ref{fig:hyperplanes}(left). We claim that this procedure is not well suited in MLC for two fundamental reasons: (1) regions defined by a collection of splitting hyperplanes are highly irregular with some regions unbounded and others with small volumes. This leads to some combinations of labels being easier to represent than others in the feature space. (2) Logistic regression does not promote feature vectors to be similar for samples that share the same label. Instead it only enforces that samples fall on the correct side of the hyperplanes.

To address these issues, we introduce a new Affine Subspace multi-label classifier (\textbf{AS-MLC}). Instead of pushing points toward different regions, our method pulls points towards different affine subspaces. This simple idea solves the two aforementioned problems. First, all affine subspaces are homogenenous in dimension such that no combination of labels is easier to represent than others. Second, pulling points towards affine subspaces makes them share similarities in the feature space, that is, the distance to the subspace.

Formally, for each label~$i$, we define two parallel $(d-e)$-dimensional affine subspaces~$(\W_i, \b_{i0})$ and~$(\W_i, \b_{i1})$, determined by the intersection of $e$~hyperplanes, where $\W_i\in\real^{e\times{}d}$ are the shared hyperplane normals and $\b_{i0}, \b_{i1}\in\real^e$~are the bias terms of both subspaces. For a given label~$i$, points with~$y_i=0$ will be pulled towards~$(\W_i, \b_{i0})$ and points with~$y_i=1$ will be pulled towards~$(\W_i, \b_{i1})$.
\newline\newline
\noindent
{\bf Training:} To train our method, we first minimize the distances of samples to their corresponding subspaces, using the following loss function term,
\begin{equation}
    \ell_1(\x, \y) = \sum_{i=1}^n \alpha_{i,y_{i}}\left\|\W_i\z + \b_{i,y_i}\right\|_2^2,
\end{equation}
where $\z=h(\x)$ and $\alpha$ is a class-label specific weight. At the same time, we also want  subspaces corresponding to the same label to be as far apart as possible from each other. This can be formalized with the additional loss term,
\begin{equation}
    \ell_2 = \sum_{i=1}^n \dfrac{1}{\|\b_{i0}-\b_{i1}\|_2^2 + \epsilon},
\end{equation}
that maximizes the distance between the parallel subspaces. Finally, to avoid that the loss terms are minimized by scaling down the magnitude of the weights, we add a regularization term to enforce that normals have unit magnitude,
\begin{equation}
    \ell_3 = \sum_{i=1}^n \tr\left|\W_i\W_i^T - \mathbf{I}\right|,
\end{equation}
where $|\cdot|$~is the element-wise absolute value, $\mathbf{I}$ is the identity matrix, and $\tr$~is the matrix trace.
Given a training dataset $\{\x^{(k)}, \y^{(k)}\}_{k=1}^K$ of images and their corresponding labels, the training procedure minimizes the weighted sum of these three terms:
\begin{equation}
    \argmin_{\theta, \phi} \dfrac{1}{K}\sum_{k=1}^K \ell_1(\x^{(k)}, \y^{(k)}) + \beta\ell_2 + \ell_3,
\end{equation}
where $\theta$~are the parameters of the feature extractor~$h$, $\beta$ is a distance weighting hyperparameter and $\phi=\{(\W_i, \b_{i0}, \b_{i1})\}_{i=1}^n$ are the weights and bias terms of our \textbf{AS-MLC}.
This loss function is trainable in an end-to-end manner.
After training, the intersections of the $2n$~learned subspaces define $2^n$~$(d-n\cdot{}e)$-dimensional affine subspaces, one for each combination of labels. See Fig.~\ref{fig:hyperplanes}(right) for an example where $n=d=2$ and~$e=1$. In this case, the final subspaces are $0$-dimensional, namely, points.
\newline\newline
\noindent
{\bf Inference: } At test time, one could use the ratio of distances to each subspace as our criterion for assigning the probability of every label (we denote this method as \textbf{AS-MLC-Distance}). However, we found that a data-driven approach reaches better performance in practice. For each label~$i$ and class~$j$, we thus build a kernel density estimation of the likelihood using the projected training data,
\begin{equation}
    p(\W_i\z \mid \y_i=j) = \dfrac{1}{K}\sum_{k=1}^K G_{\delta}\left(\W_i(\z - \z^{(k)})\right),
\end{equation}
where $G_{\delta}$~is the Gaussian kernel with bandwidth~$\delta$, $\z^{(k)}=h(\x^{(k)})$~is the descriptor vector of the $k$-th element of the training data, and $\z=h(\x)$~is the descriptor vector of the input image. Note that bias terms are not required to define the density, as they are implicitly encoded in the set of descriptor vectors~$\{\z^{(k)}\}_{k=1}^K$. We define the posterior (assuming uniform priors)

\begin{equation}
    g(\z)_i \equiv P(\y_i=1\mid \W_i\z) =
    \dfrac{p(\W_i\z \mid \y_i=1)}
    {\sum_{j\in\{0,1\}} p(\W_i\z \mid \y_i=j)},
\end{equation}
which is the $i$-th output of our multi-label binary classifier~$g$.

% !TEX root = top.tex
% !TEX spellcheck = en-US

\section{Experiments}
To evaluate the performance of our proposed method, we perform experiments on two medical MLC image datasets. %Our goal is to verify that our proposed \textbf{AS-MLC} method increases the performance compared to other common approaches.

\subsection*{Dataset 1 -- OCT Biomarker identification}
This dataset consists of volumetric Optical Coherence Tomography (OCT) scans of the retina with 11 pathological biomarker labels annotated. The data is split into 23'030 and 1'029 images for the training and testing sets, respectively, with no patient images in both splits. The image labels include: Healthy, Sub Retinal Fluid, Intraretinal Fluid, Intraretinal Cysts, Hyperreflective Foci, Drusen, Reticular Pseudodrusen, Epirential Membrane, Geographic Atrophy, Outer Retinal Atrophy and Fibrovascular PED. Fig.~\ref{fig:examples} (Left) shows a training example with two biomarkers being present.

To compare our approach to existing methods, we evaluted a number of baselines using two different NN architectures: a pre-trained DRND-54 \cite{Yu2017} and %(Imagenet Accuracy: 78.8\%)
a ResNet-50 \cite{He2016-tq}. %(Imagenet Accuracy: 76\%) networks.
All methods are trained using the Adam optimizer~\cite{Kingma2014-uu} with a base learning rate of $10^{-3}$. We apply the same data augmentation scheme (flipping, rotation, translation, gamma and brightness) for all experiments. Results are reported with 5-fold cross validation where the training data was split into training 80\% and validation 20\%. Baselines include:
\begin{itemize}
\item {\bf Softmax:} Two class outputs per label that are normalized using the softmax operator and the binary cross-entropy loss is optimized.
\item {\bf Ranking:} We use the ranking loss as described by Li et al. \cite{Li2017-if}. As ranking losses are typically thresholded, we omit this threshold and scale outputs between 0 and 1 during training and testing. We acknowledge that this is a disadvantage for the ranking method, but include it for the sake of comparision.
\item {\bf Ml-kNN:} We apply a distance weighted kNN (n=50) to $\bf{z}$ extracted from the {\bf Softmax} method as in~\cite{Zhang2007-cu}.
\item {\bf AS-MLC:}  We set $\beta=5,\alpha=1$ and $e=32$. The Gaussian kernel densitiy estimation bandwidth is set to $\delta=0.1$ and uses the features of the training images and their horizontally flipped versions. We also compare to the distance function method \textbf{AS-MLC-Distance}.
\end{itemize}

\begin{table}[t!]
\sisetup{
table-alignment = center,
table-column-width= 0.14\textwidth,
table-figures-integer = 5,
table-figures-decimal = 0,
}
\centering
\resizebox{\textwidth}{!}{\begin{tabular}{lccccc}
\toprule
\textbf{} & \textbf{Loss} & \textbf{Macro mAP} & \textbf{Micro mAP} & \textbf{Macro mAP*} & \textbf{Micro mAP*} \\ \midrule
\parbox[t]{4mm}{\multirow{4}{*}{\rotatebox[origin=c]{90}{ResNet-50 }}}
& Softmax & 0.790 & 0.805 & 0.797 & 0.815 \\
& Ranking & 0.747 & 0.759 & 0.764 & 0.752\\
& Ml-kNN & 0.770 & 0.799 & 0.781 & 0.806 	\\
& AS-MLC-Distance & 0.762  &  0.778 & 0.773 & 0.789  \\
& AS-MLC & \bf{0.800} &	\bf{0.814} & \bf{0.810}  & \bf{0.822} \\
\bottomrule
\parbox[t]{3mm}{\multirow{5}{*}{\rotatebox[origin=c]{90}{DRND-54}}}
& Softmax & 0.806	& 0.801 & 0.813 & 0.830 \\
& Ranking  & 0.770 &	0.782 & 0.781 & 0.789 \\
& Ml-kNN & 0.790 &	0.817 & 0.805 &	0.828 \\
& AS-MLC-Distance & 0.824 & 0.814 & 0.823 & 0.834  \\
& AS-MLC & \bf{0.831} &	\bf{0.848} & \bf{0.840} & \bf{0.8567} \\
\bottomrule
\end{tabular}}
\caption{Experimental results comparing our proposed method to other approaches. Testing is performed by taking the weights at the epoch where the maximum macro mAP was achieved on the validation set. (*) indicates test time augmentation.}
\label{table:table-results-obc}
\end{table}

\begin{figure}[!t]
    \centering
       \includegraphics[width=0.49\textwidth]{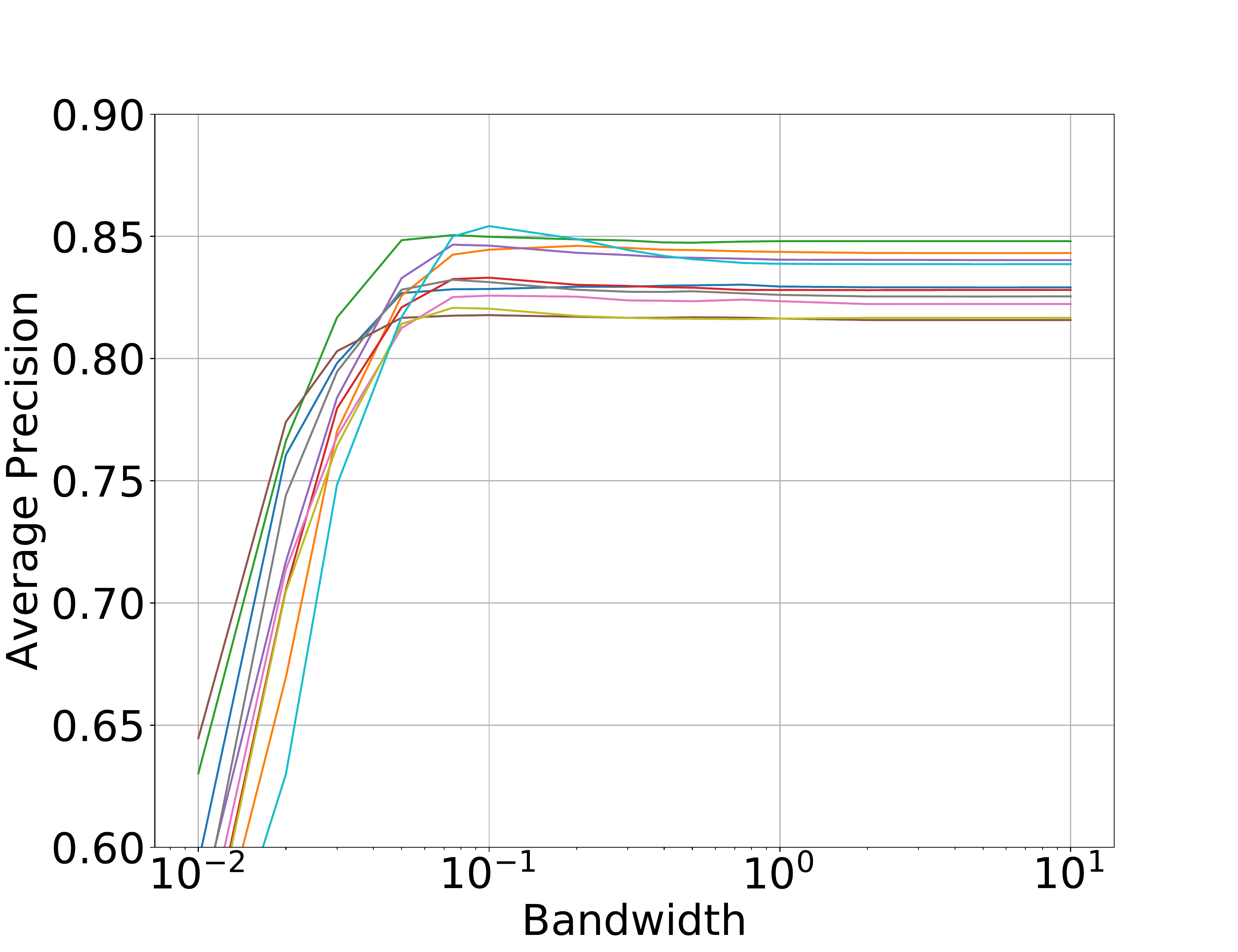}
      \includegraphics[width=0.49\textwidth]{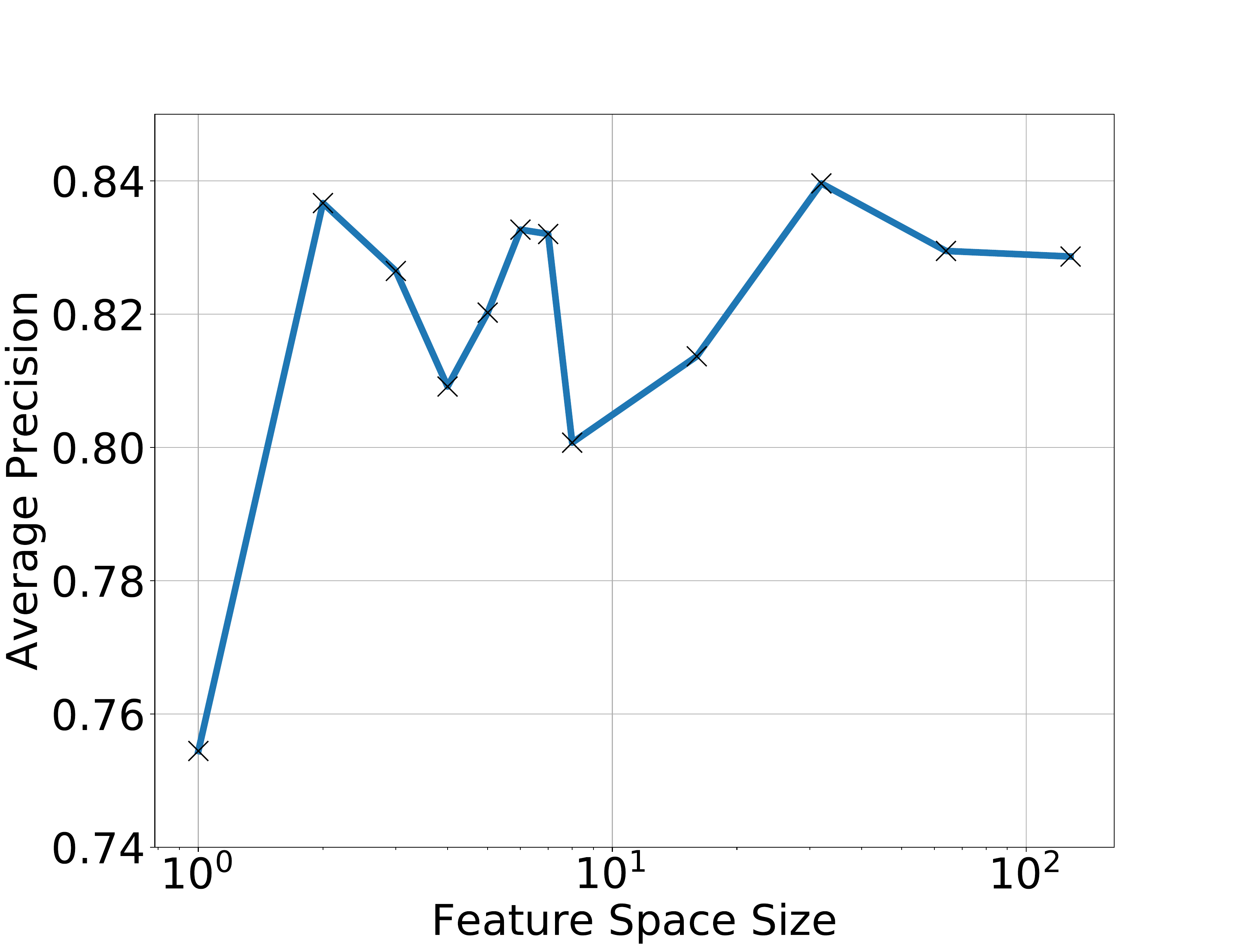}
\caption{\textbf{Left:} 10 fold cross validation of bandwith size using a fixed feature size $e=32$. \textbf{Right:} Impact analysis for the feature dimension size $e$ at a fixed bandwidth $\delta=0.1$.}
\label{fig:bandwidth_dimensions}
\end{figure}

\noindent
Mean Average Precision (mAP) results are presented in Table \ref{table:table-results-obc} and show that our proposed method outperforms the commonly used loss functions for all metrics and both networks. We show both the micro and macro averaged results, when using and not using test time data augmentation (original image + left/right flip).
We see a performance increase of up to 5.7\% using our method over the softmax cross-entropy loss.
In Fig.~\ref{fig:bandwidth_dimensions} (Left), we show the influence of the bandwidth $\delta$ value when using a 10-fold cross-validations. Here we see that unless the bandwidth value is chosen to be too small, the performance remains stable for a wide range of values. Similarly, we also analyse the effects of the size of the feature space $e$, which we consider to be an additional hyperparameter. From Fig.~\ref{fig:bandwidth_dimensions} (Right) we can conclude that extremely small feature space sizes are not sufficient for our method but for values greater than 5, performances are consistently high.

\subsection*{Dataset 2 -- ChestX-ray14}
The dataset contains 112'120 X-ray scans associated to 14 different labels \cite{Wang2017-zo}.
The data is split according to the original patient-level data splits which results in 70\% training, 10\% validation and 20\% test sets.
We resize images to 512x512 pixels and optimize the network using the Adam optimizer \cite{Kingma2014-uu}.
In this case, we compare the weighted cross-entropy loss and add the weighting term $\alpha$ to the class labels due to the significant class imbalances in the data (\ie weights are equal to the inverse class occurence). For all experiments we use a DRND-54~\cite{Yu2017} CNN as the base architecture. We fix the hyperparamters $\beta=5$ and find the best bandwidth using cross-validation ($\delta=1.0$).
\begin{table}[t!]
\sisetup{
table-alignment = center,
table-column-width= 0.14\textwidth,
table-figures-integer = 5,
table-figures-decimal = 0,
}
\centering
\begin{tabular}{c|ccccccccc}
\toprule
\textbf{Method} & \textbf{Atelectasis} & \textbf{Cardiomegaly}  & \textbf{Effusion} & \textbf{Infiltration} & \textbf{Mass}   &  \\ \midrule
Original~\cite{Wang2017-zo} & 0.7003 & 0.8100 & 0.7585 & 0.6614 & 0.6933\\
Softmax &  0.7290 & \textbf{0.8514} & 0.7893 & \textbf{0.6692} & 0.7853\\
AS-MLC &  \textbf{0.7471} & 0.8481 & \textbf{0.8203} & 0.6647 & \textbf{0.7957}\\

\midrule
\textbf{Method} & \textbf{Nodule} & \textbf{Pneumonia}  & \textbf{Pneumothorax} & \textbf{Consolidation}  & \textbf{Edema}   \\ \midrule
Original~\cite{Wang2017-zo} &  0.6687 & 0.658 & 0.7993 & 0.7032 & 0.8052 \\
Softmax & 0.7217 & \textbf{0.700} & 0.8371 & 0.7134 & 0.8291\\
AS-MLC & \textbf{0.7759} & 0.6997 & \textbf{0.8663} & \textbf{0.7294} & \textbf{0.8306}\\

\midrule
\textbf{Method} & \textbf{Emphysema} & \textbf{Fibrosis}  & \textbf{PT} & \textbf{Hernia} & \textbf{Average}  \\ \midrule
Original~\cite{Wang2017-zo} & 0.833 & 0.7859 & 0.6835 & 0.8717 &  0.7451\\
Softmax & 0.9051  & 0.8042 & 0.7283 & 0.8560 & 0.7800\\
AS-MLC & \textbf{0.9200} & \textbf{0.8222} & \textbf{0.7626} & \textbf{0.9288 }& \bf{0.8008}\\
% \bottomrule
\end{tabular}
\caption{Area under the curve values for ROC results on the ChestX-ray14 dataset.}
\label{table:table-results}
\end{table}

Results of our {\bf AS-MLC} method are given in Table \ref{table:table-results} and yield a 0.8008 mean AUC.
Our method thus outperforms the softmax cross entropy loss by nearly 3\% in mean ROC values.
These results using a standard network are in range of previously published state-of-the-art results that used large amounts of additional training data~\cite{Guendel2018-wx} (0.806), attention based models~\cite{Tang2018-jm,Guan2018-yq} (0.8027 and 0.816) and significantly outperform the original publication~\cite{Wang2017-zo}.

% % \textbf{Label} & \textbf{Original \cite{Wang2017-zo}}  & \textbf{Softmax} & \textbf{Proposed}  \\ \midrule
% Atelectasis & 0.7003 &  0.7290 & 0.7471  \\
% Cardiomegaly & 0.81 &  0.8514 & 0.8481 \\
% Effusion & 0.7585 &  0.7893 & 0.8203 \\
% Infiltration & 0.6614 &  0.6692 & 0.6647\\
% Mass & 0.6933 & 0.7853 & 0.7957 \\
% Nodule & 0.6687 & 0.7217 & 0.7759  \\
% Pneumonia & 0.658 & 0.700 & 0.6997 \\
% Pneumothorax & 0.7993  & 0.8371 & 0.8663\\
% Consolidation & 0.7032  & 0.7134 & 0.7294 \\
% Edema & 0.8052  & 0.8291 & 0.8306  \\
% Emphysema & 0.833 & 0.9051 & 0.9200 \\
% Fibrosis & 0.7859 & 0.8042 &  0.8222\\
% PT & 0.6835 & 0.7283 & 0.7626 \\
% Hernia & 0.8717 & 0.856 & 0.9288 \\
% \midrule
% Average & 0.7451  & 0.780 & \bf{0.8008}\\

% !TEX root = top.tex
% !TEX spellcheck = en-US

\section{Conclusion}
\label{sec:conc}
We presented \textbf{AS-MLC}, a novel MLC method which attempts to overcome the short-comingss of classical MLC methods by classifying in affine subspaces.
To do so, we propose a novel loss function which pulls class-labels towards affine subspaces and maximizes their distance.
We evaluated our method on two datasets and showed that it consistently outperforms state-of-the-art approaches.
The proposed method is a plug-in replacement for standard deep learning architectures and can be learnt end-to-end using standard backpropagation.
In the future we wish to investigate how to extract attention maps from the predictions as the application of methods such as GradCAM~\cite{Selvaraju2016-jg} are no longer directly applicable.

\section*{Acknowledgements}
This work received partial financial support from the Innosuisse Grant \#6362.1 PFLS-LS.

{
\small
\bibliographystyle{splncs}
%\bibliography{string,vision,learning,biomed,optim}
\bibliography{paper1551_top}
}

\end{document}